\RequirePackage{fix-cm}

\documentclass[twocolumn]{svjour3}          
\usepackage{microtype}
\smartqed  
\usepackage{graphicx}
\usepackage[caption=false,font=normalsize,labelfont=sf,textfont=sf]{subfig}
\usepackage{amsmath,amsfonts}
\usepackage{url}
\usepackage{xcolor}
\usepackage{microtype}
\definecolor{highlight}{rgb}{0, 0, 0}

\journalname{Journal of Real-Time Image Processing}
\begin{document}
\hyphenation{self-super-vis-ed under-explored sur-veil-lan-ce super-vis-ed}
\title{SelfReDepth
}
\subtitle{Self-Supervised Real-Time Depth Restoration for Consumer-Grade Sensors}

\titlerunning{}        

\author{
Alexandre Duarte \and
Francisco Fernandes \and
Jo\~ao M. Pereira \and
Catarina Moreira \and
Jacinto C. Nascimento \and
Joaquim Jorge
}


\institute{
Alexandre Duarte \textsuperscript{1} \at
\email{alexandre.a.duarte@tecnico.ulisboa.pt}
\and
Francisco Fernandes \textsuperscript{2} \at
\email{francisco.fernandes@tecnico.ulisboa.pt}
\and
Jo\~ao M. Pereira \textsuperscript{1,2} \at
\email{jap@inesc-id.pt}
\and
Catarina Moreira \textsuperscript{2,4} \at
\email{catarina.pintomoreira@uts.edu.au}
\and
Jacinto C. Nascimento \textsuperscript{1,3} \at
\email{jacinto.nascimento@tecnico.ulisboa.pt}
\and
Joaquim Jorge \textsuperscript{1,2} \at
\email{jaj@inesc-id.pt}
\and
\textsuperscript{1} Instituto Superior T\'ecnico, Universidade de Lisboa (IST-UL), Lisbon, 1000-029, Portugal.\\
\textsuperscript{2} Instituto de Engenharia de Sistemas e Computadores, Investiga\c c\~ao e Desenvolvimento (INESC-ID), Lisbon, 1000-029 Portugal.\\
\textsuperscript{3} Institute for System and Robotics (ISR), Instituto Superior T\'ecnico, Universidade de Lisboa (IST-UL), Lisbon, 1049-001, Portugal.\\
\textsuperscript{4} Human Technology Institute, University of Technology Sydney, Sydney, Australia.
}

\date{Received: September 14, 2023 / Accepted: June 3, 2024}

\maketitle

\begin{abstract}
Depth maps produced by consumer-grade sensors suffer from inaccurate measurements and missing data from either system or scene-specific sources. Data-driven denoising algorithms can mitigate such problems, 
\textcolor{highlight}{However, they require vast amounts of ground truth depth data}. Recent research has tackled this limitation using self-supervised learning techniques, but it requires multiple RGB-D sensors. Moreover, most existing approaches focus on denoising single isolated depth maps or specific subjects of interest \textcolor{highlight}{highlighting a need 
for methods that can effectively denoise depth maps in real-time dynamic environments}.
This paper extends state-of-the-art approaches for depth-denoising commodity depth devices, proposing SelfReDepth, a self-supervised deep learning technique for depth
\textcolor{highlight}{restoration, via denoising and hole-filling by inpainting} of full-depth maps captured with RGB-D sensors. The algorithm targets depth data in video streams, utilizing multiple sequential depth frames coupled with color data to achieve high-quality depth videos with temporal coherence. Finally, SelfReDepth is designed to be compatible with various RGB-D sensors and usable in real-time scenarios as a pre-processing step before applying other depth-dependent algorithms.
Our results demonstrate 
our approach's real-time performance on real-world datasets shows that it outperforms state-of-the-art methods in denoising and restoration performance at over 30fps on Commercial Depth Cameras, with potential benefits for augmented and mixed-reality applications.
\end{abstract}

\section{Introduction}

Depth information is pivotal in many applications, from digital entertainment to virtual and augmented reality~\cite{jorge19}. It is the backbone for digital object and environment modeling~\cite{newcombe2011kinectfusion,Matterport3D} and cost-effective motion capture solutions~\cite{gao2015leveraging}.

Pose estimation derived from depth data finds utility in diverse fields such as physiotherapy~\cite{gabel2012full,capecci2016physical}, video surveillance~\cite{zhang2012water,liu2015detecting}, and human-computer interaction~\cite{ren2011robust}. Depth data also aids autonomous navigation~\cite{feng2018towards} and enhances security measures through facial recognition~\cite{oyedotun2017facial}.

Consumer depth devices, often employing low-cost LiDAR, Structured Light, or Time-of-Flight technologies, are instrumental in these applications. Among these, the Microsoft Kinect v2 stands out for its balance of quality, availability, and affordability. However, consumer-grade sensors like Kinect v2 still grapple with noisy and incomplete data issues.

Efforts to address these quality issues span traditional smoothing techniques to data-driven machine learning algorithms. Many adopt supervised learning with neural networks, training models on noisy-clean data pairs $(\hat{x}, y)$ to minimize empirical risk.

However, acquiring clean training data is non-trivial. Recent attention has thus shifted towards self-supervised techniques, such as Noise2Noise~\cite{lehtinen2018noise2noise}, which leverages noisy-noisy data pairs $(\hat{x}, \hat{y})$ for training, and minimizing the cost function  $g(\theta) = 
{\mathrm{argmin}}_\theta \sum_i^N {\cal L} \left(f_\theta \left( \hat{x}_i \right), \hat{y}_i \right)$, where the network $f_\theta \left( \hat{x}_i \right)$ is parameterize by $\theta$.

Despite their efficacy in various domains, self-super\-vis\-ed methods for depth data restoration remain underexplored, largely due to the intricate noise patterns in consumer-grade sensors.

Our paper introduces \textbf{SelfReDepth} (\textbf{SReD}), a novel self-supervised, real-time depth data restoration technique optimized for the Kinect v2. \textcolor{highlight}{SelfReDepth introduces a convolutional autoencoder architecture inspired by U-Net, specifically designed to process sequential depth frames efficiently. This design choice directly responds to the need for maintaining temporal coherence in dynamic scenes, a gap often left unaddressed by traditional single-frame denoising approaches. Furthermore, SelfReDepth incorporates RGB data into the depth restoration process as an innovative way to enhance the accuracy of inpainting missing pixels by providing contextual color information. This method significantly improves the restoration quality by providing additional context that depth data alone lacks.} 
Our contributions are fourfold:
\textbf{(1)} We employ a convolutional autoencoder with an architecture akin to U-Net~\cite{ronneberger2015unet} to process sequential frames.
\textbf{(2)} Our method achieves real-time performance and temporal coherence by adopting a video-centric approach.
\textbf{(3)} We incorporate RGB data to guide an inpainting algorithm during training, enhancing the model's ability to complete missing depth pixels.
\textbf{(4)} Our approach maintains a 30fps real-time rate while outperforming state-of-the-art techniques.


\section{Background and Related Work}
\label{sec:back_rel}

In recent years, depth-sensing technology has emerged as a pivotal tool in various applications, from gaming to augmented reality and robotics. The promise of capturing the third dimension, depth, has opened up new horizons in computer vision, augmented reality, and human-computer interaction.
Next, we introduce some concepts and methodologies related to the present work.

\begin{figure*}[t]
    \centering
    \subfloat[training]{
        \label{figModel:train}
        \includegraphics[width=0.45\linewidth]{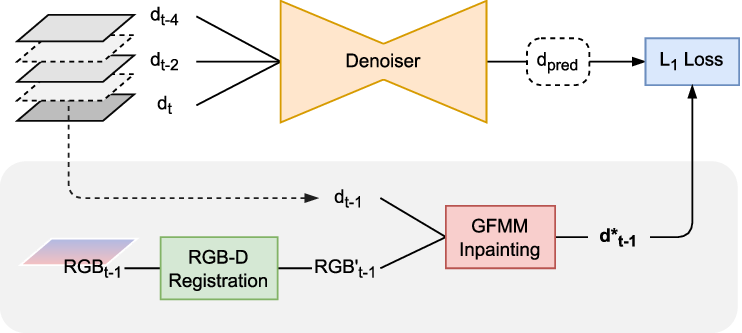}} \quad
    \subfloat[Inference]{
        \label{figModel:inf}
        \raisebox{40pt}{\includegraphics[width=0.3\linewidth]{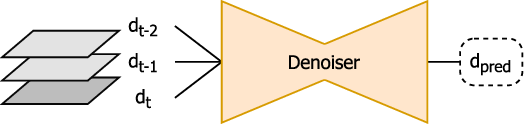}}}
    \caption{Full overview of SelfReDepth's architecture.}
    \label{figModel}
\end{figure*}

{\color{highlight}
\vspace{10pt}\noindent\textbf{Denoising vs. inpainting}: The distinction between {\it denoising} and {\it inpainting} is important to be stressed, as these terms will be used throughout this work constituting important stages of the proposed methodology. Denoising and inpainting are two core image processing problems. As the name suggests, denoising removes noise from an observed noisy image, while
inpainting aims to estimate missing image pixels. Both denoising and inpainting are inverse
problems: the common goal is to infer an underlying image
from incomplete/imperfect observations.
Formally, in both problems the observed image $\bf Y\in \mathbb{R}^{M'\times N'}$ is modeled as 
    ${\bf Y}= {\cal F}({\bf X}) + \eta \label{eq:den-paint}$
where ${\bf X}\in \mathbb{R^{M\times N}}$ is the unknown (original) image and $\eta$ is the 
observed noise.
The difference between the denoising and the inpainting emerges from the mapping 
${\cal F}: \mathbb{R}^{M\times N} \mapsto \mathbb{R}^{M'\times N'}$ that expresses a linear degradation operator that could represent a convolution
or a masking process. 
In concrete, the denoising process means that ${\cal F}$ is an {\it identity projector}, having ${\cal F}={\bf H}$, such that  ${\bf H}=\bf I$,  (with $\bf I$ the identity matrix). In the inpainting process, ${\cal F}$ is a {\it selection operator}. In practice, this corresponds to having the same $\bf H$ as before. However, this matrix only contains a subset of the rows of ${\bf I}$, accounting for the loss of pixels. {\color{highlight}

Having formally defined the two concepts, we stress that both are used in restoration problems, as we propose in this work. In this paper, two major contributions are offered for restoration, concretely: $(i)$ a new denoising method. Contrasting with Noise2Noise~\cite{lehtinen2018noise2noise} that applies denoising in traditional images, we extend the  framework to depth images that require a new learning strategy to handle depth information 
(see top branch in Fig.~\ref{figModel} and Sec.~\ref{sec:target-generation}), and $(ii)$ inpainting approach where we integrate a new two-stage pipeline comprising an RGB-Depth registration and a Fast Marching Method stage (see a bottom branch in Fig.~\ref{figModel}).
}
}

\begin{figure*}[!h]
    \centering
    \subfloat[training]{
        \includegraphics[width=0.45\linewidth]{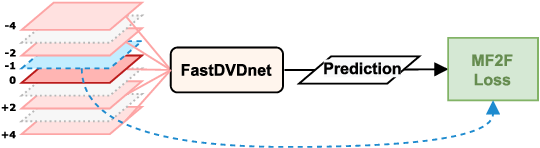}}
    \hfill
    \subfloat[inference]{
        \includegraphics[width=0.38\linewidth]{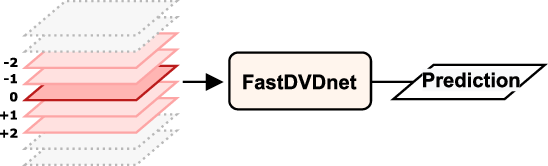}}
    \caption{Multi-Frame-to-Frame (MF2F)~\cite{dewil2021self} architecture with distinct strategies from the training and inference steps.}
    \label{figMF2F}
\end{figure*}

\vspace{10pt}\noindent\textbf{Problems with low-quality depth}: Despite all the progress made in in-depth sensing hardware for consumer devices, depth cameras (such as the Kinect v2) still suffer from many of the same problems that previous iterations also did, namely, noisy measurements and depth holes
\cite{mallick2014characterizations}.

These depth holes typical of Time-Of-Flight (ToF) 
devices have multiple causes~\cite{jiang2018kinect,song2017depth} including: 
    \textbf{(1)} Measuring regions that are outside the distance range of the sensor,
    \textbf{(2)} Highly reflective objects in the scene,
    \textbf{(3)} Measurements near the edges of the camera's field-of-view (FoV). 
Smaller holes, on the other hand, can appear in one of two types:
    \textbf{(1)} Isolated points caused by physical and lighting interferences on the sensor,
    \textbf{(2)} Thin outlines around objects due to the scattering of infrared rays at shallow angles and sharp edges.

Besides missing depth values, the measurement inaccuracy in the successfully captured points is a concerning issue, leading to noisy depth maps. The noise produced by Kinect v2 is significantly less severe than in the previous generation. Nevertheless, it is still very much present. Experimental analysis~\cite{zennaro2015performance,lachat2015assessment,tolgyessy2021evaluation} has shown that there is a direct correlation between the noise observed in Kinect v2's depth maps and various physical factors, including distance, angle, material, color, warm-up time, to quote a few. Furthermore, there is a general consensus~\cite{dai2019new,kweon2014noise,basso2018robust} that this noise can be described as the sum of two different sources of noise:
    (1) Random noise, associated with pixel-based local distortions caused by physical factors like color and others mentioned above,
    (2) Systematic bias, associated with the wiggling error that radially increases as the measurements get closer to the edges of the sensor's FoV.


\vspace{10pt}\noindent\textbf{Deterministic denoisers}, or manual
denoising algorithms that do not rely on machine learning 
were
the first noise reduction 
techniques to be developed targeting depth data
\cite{essmaeel2012temporal}.
These can generally be divided into three main categories: (i) {\em filter-based} denoisers, (ii) {\em outlier removal} techniques, and (iii) {\em calibration} methods.

Filter-based denoisers work by applying smoothing and sharpening filters, such as bilateral filters~\cite{maimone2012enhanced,zhang2008adaptive}, joint bilateral filters~\cite{chen2012depth,chaudhary2016approach}, anisotropic filters~\cite{liu2016computationally} and zero block filters~\cite{liu2016computationally}, to leverage spatial pixel neighborhoods through sliding pixel windows ({\em i.e.} kernels). The preservation of edge sharpness is particularly difficult to achieve using filters; thus, some denoisers introduce specialized techniques, such as RGB-D alignment~\cite{chen2012depth} and contextual image partitioning~\cite{chaudhary2016approach}.

Other works, instead, focus exclusively on removing incorrect or low-quality depth points rather than correcting them. This can be seen, for instance, in \cite{dai2019new} for cleaning hand depth scans and in \cite{wan2020edge} for cleaning body scans, later merged to form a complete body point cloud.
Finally, some works tackle the denoising problem from a calibration standpoint, focusing on alleviating systematic errors affecting consumer-grade sensors by fitting planes or splines to the raw measurements~\cite{kweon2014noise,lachat2015assessment} or generating specialized noise correction maps~\cite{basso2018robust}.


\vspace{10pt}\noindent\textbf{Self-supervised Denoisers:} Noise2Noise~\cite{lehtinen2018noise2noise} pioneered self-supervised image denoising, showing that a denoising model trained with only noisy data can achieve quality results on par with supervised learning strategies.
The shift in the learning method from supervision to self-supervision resides primarily in the training data.
Specifically, Noise2Noise~\cite{lehtinen2018noise2noise} uses input-target pairs of the form $\left(\hat{x},\hat{y}\right)=\left(x+n_1,x+n_2\right)$, where $x$ is the base signal (the undamaged data that we want to uncover) and $n_1$ and $n_2$ are two independent noise instances following the same statistical distribution. From the above, the  Noise2Noise strategy differs from the noise-clean data pairs $\left(\hat{x},y\right)$ used in supervised learning and has the advantage of not requiring clean target images. 

Nonetheless, Noise2Noise~\cite{lehtinen2018noise2noise} has some data limitations, encouraging subsequent works to propose further improvements. Towards this challenge, \cite{calvarons2021improved} proposes two data permutation techniques to increase the number of noisy training pairs. On the other hand, Noise2Void~\cite{krull2019noise2void} eliminates the need for quasi-similar input-target pairs (not always easy to obtain) by training the model to predict a central pixel using noisy-void training pairs $\left(\hat{x},-\right)$ and a blind-spot mask to avoid learning the identity. Noise2Self~\cite{batson2019noise2self} later expanded on this by proposing a more generalized model.


Going further, some works, namely
Probabilistic Noise2Void~\cite{krull2020probabilistic},
SURE~\cite{metzler2018unsupervised},
Noisier2Noise~\cite{moran2020noisier2noise} and NoiseBreaker~\cite{lemarchand2020noisebreaker}, managed to improve denoising for specific 
distributions.
On the other hand, GAN2GAN~\cite{cha2019gan2gan} combines a generative model, Self2Self~\cite{quan2020self2self} introduces training with a single data sample, and GainTuning~\cite{mohan2021adaptive} proposes an ever-adapting model.


\vspace{10pt}\noindent\textbf{Spatio-temporal Denoisers} 
are an extension of image denoising where the coherence of temporal locality is also considered to provide visual continuity in the final denoised videos. Likewise, clean data may also not be easy to obtain for this task, and thus, blind training comes with great interest. A simple multi-frame self-supervised strategy for denoising can be extrapolated from Noise2Stack~\cite{papkov2021noise2stack}, in which a self-supervised approach is proposed to denoise MRI 
data using adjacent 
sets from a stack of layered MRI brain scans.

Self-supervised denoising techniques targeting color videos have also been developed, using the multi-frame input concept described in Noise2Stack~\cite{papkov2021noise2stack} combined with additional self-supervision techniques. Multi-Frame-to-Frame (MF2F)~\cite{dewil2021self}
(see Fig.~\ref{figMF2F})
takes the FastDVDNet~\cite{tassano2020fastdvdnet} supervised video denoising network, composed by cascaded U-Net~\cite{ronneberger2015unet} autoencoders, and applies its own self-supervised loss.


Similarly, UDVD~\cite{sheth2021unsupervised} uses a cascaded structure akin to FastDVDNet~\cite{tassano2020fastdvdnet} but performs the network pass 4-fold, each with the input frames at a different rotation (0º, 90º, 180º and 270º). The four outputs generated, one for each of the four rotations, are then rotated back to 0 degrees and combined to form the final production.

\noindent\textbf{Depth Completion:} Alongside inaccurate data points, low-quality depth maps also suffer from missing or invalid data. Depth completion, also known as hole-filling, is a well-known and vastly researched area that falls under the umbrella of image inpainting~\cite{yu2018generative,liu2018image,xiong2019foreground}.

The effect of using Noise2Noise and similar algorithms over images with missing data without any prior inpainting is that the majority of depth holes remain untreated in the final images, and even in methods that deal with multiple consecutive frames, there is insufficient data to fill these gaps in most cases.

As in-depth denoising, depth completion has been approached using traditional and deep neural methodologies. Traditional techniques typically rely on either filtering algorithms, which classify the holes and apply dedicated filters, such as PDJB, DJBF, and FCRN, or boundary-extending algorithms, based on FMM or the Navier-Stokes equation~\cite{bertalmio2001navier}.

The Fast Marching Method (FMM) inpainting, in particular, was originally proposed for color image inpainting~\cite{telea2004image} and works by progressively shrinking the boundaries of hole regions inwards, until all pixels have been filled, using the equation
\begin{equation}
\label{FMM_original_full_eq}
    I(p) = \dfrac{\sum \limits_{q \in N(p)} {w(p, q) \cdot \left[I(q) + \nabla I(q) \cdot (p-q)\right]}}
            {\sum \limits_{q \in  N(p)} {w(p, q)}}
\end{equation}
where $p$ is the pixel being inpainted, $N(p)$ is a neighborhood 
pixels of $p$, $w(p,q)$ is a function that determines how much pixel $q$ contributes to the inpainting of 
$p$, $I$, and $\nabla I$ represent the image and discrete gradient of the image,  respectively. This algorithm was 
extended to depth completion, introducing improvements like the use of aligned color as a guiding factor for the weight function and to define the order of computations~\cite{liu2012guided}, and the use of a pixel-wise confidence factor~\cite{li2021depth}.

Sparse depth maps, generally captured with LiDAR sensors, suffer especially from large patches of missing depth data and have particular time limitations, as they are commonly linked with autonomous driving. Thus, more advanced 
techniques have been developed, relying on both supervised~\cite{li2021fastcompletion} and self-supervised~\cite{ma2019self,choi2021selfdeco,feng2021advancing} deep convolutional neural networks assisted by color information to fill large depth gaps.

\section{Our Approach}
\label{sec:approach}

Building upon recent advancements in self-supervised data denoising research, the proposed SReD offers a novel approach for denoising and inpainting low-quality depth maps. It leverages the flexibility and adaptability of deep learning models while eliminating the need for reference data - a highly desirable feature also found in deterministic denoisers. SReD was designed with a specific practical use case in mind, incorporating several additional requirements during its design and development. Specifically, our technique aims to: $(i)$ denoise and restore as much of the initial depth maps as possible, $(ii)$ operate with a single RGB-D device, $(iii)$ facilitate direct sensor data streaming, and $(iv)$ strive for temporal coherence and real-time performance.

Naturally, these requirements posed challenges that influenced the architectural decisions. For example, to achieve depth video denoising with temporal coherence, it is logical to design a method that utilizes multiple sequential frames, similar to MF2F~\cite{dewil2021self}. {\color{highlight} However, given real-time constraints, only frames up to the most recent one are considered. This contrasts with  MF2F, which incorporates two subsequent frames at the cost of adding considerable lag.} Additionally, an architecture with faster inference is preferable over a more complex one to meet real-time performance criteria.

\subsection{SelfReDepth's Architecture}
\label{sec:SelfReDepth-Architecture}


SReD's architecture, particularly its neural network and learning method, takes inspiration from previous self-supervised denoisers, mainly Noise2Noise~\cite{lehtinen2018noise2noise}, MF2F~\cite{dewil2021self} and Noise2Stack~\cite{papkov2021noise2stack}, adapting their proposed denoising models to the specific case of online denoising in depth map sequences. As depicted in Fig.~\ref{figModel}, the architecture differs between its training and inference stages. The model uses a dilated input during the training stage, as proposed in MF2F~\cite{dewil2021self}. 
The autoencoder is trained with noisy pairs $(\hat{x}, \hat{y}) = ([d_{t-4},d_{t-2},d_{t}],d_{t-1})$, where $d_{t-k}$ is the depth frame at time instant $t-k$, with $k\in\{0,1,2,4\}$. This technique improves the denoising results during inference and prevents the network from learning the identity by hiding frame $d_{t-1}$ from the input. Since  $d_{t}$ and $d_{t-1}$ are consecutive time frames, it is also plausible to assume they are similar in content while having different instances of noise, 
making them a suitable image pair for the Noise2Noise-style training.

Moreover, noisy depth frames frequently have regions persistently composed of depth holes in both the input and target frames, making these regions impossible to ``denoise'' using a standalone denoising network. As such, during training, the target frame $d_{t-1}$ is inpainted with an FMM inpainting algorithm guided by the registered color frame ${\rm RGB}_{t-1}$, providing a way for the denoiser to learn how to fill depth-holes.

In summary, SReD's training architecture has two distinct main blocks, depicted in Fig.~\ref{figModel:train}: $(i)$ a denoising convolutional autoencoder with dilated input; and $(ii)$ a target generation pipeline responsible for creating inpainted targets. During inference (Fig.~\ref{figModel:inf}), the target generation is removed, contributing to faster performance and the denoiser shifts to non-dilated input ({\em i.e.}, taking the frames $[d_{t-2},d_{t-1},d_{t}]$),  estimating a denoised/inpainted instance of the frame $d_{t}$.

\subsection{Target Generation}
\label{sec:target-generation}

The denoiser requires a learning strategy to handle depth holes. To achieve this, SReD generates target frames through deterministic inpainting. This deterministic approach consists of two stages: $(i)$ computing a registered RGB image and $(ii)$ using the previous result to apply guided inpainting to the damaged depth frame. The selection of this strategy rests on three primary reasons:
(1) The prevalence of depth holes in general consumer depth data is sufficiently low for a deterministic approach to yield acceptable inpainting results.
(2) It avoids the need for reference data.
(3) From a temporal performance perspective, it only introduces computational time during training.


\vspace{10pt}\noindent\textbf{RGB-D Registration}: 
RGB-D devices collect color and depth with physically separate sensors/cameras, and often also different resolutions and FoV. Therefore, aligning the RGB and depth frames simultaneously captured must be done with a registration algorithm and requires acquiring the extrinsic and intrinsic parameters of the device, namely:
\begin{itemize}
\item Focal length $f_d = \begin{bmatrix} f_{d,x} & f_{d,y} \end{bmatrix}^\top$ and principal point $c_d = \begin{bmatrix} c_{d,x} & c_{d,y} \end{bmatrix}^\top$ of the depth/IR sensor,
\item Focal length $f_{rgb} = \begin{bmatrix} f_{rgb,x} & f_{rgb,y} \end{bmatrix}^\top$ and principal point $c_{rgb} = \begin{bmatrix} c_{rgb,x} & c_{rgb,y} \end{bmatrix}^\top$ of the RGB sensor,
\item Rotation matrix $R$, which encodes the rotation from the RGB sensor view to the depth/IR sensor,
\item Translation vector $T$ translates from the RGB sensor's position to the depth/IR sensor's position.
\end{itemize}

Following \cite{zhou2012study}, RGB-D registration is performed through a series of coordinate transformations that map depth values captured from the depth sensor's point-of-view to color values in the RGB camera's point-of-view. To achieve this, the depth data, given as a depth map, is first converted to a point format where for each pixel coordinate $\begin{bmatrix} x_d & y_d \end{bmatrix}^\top$ exists a 3-d point $X_d = \begin{bmatrix} x_d & y_d & z_d \end{bmatrix}^\top$ with $z_d = \mathrm{depth}(x_d, y_d)$, and then transformed from \textit{Depth Image Coordinate Space} to \textit{RGB Image Coordinate Space}, $X_{rgb}$, using the following equalities:
\begin{equation}
\label{reg_depth_cam}
    X'_d = \begin{bmatrix} x'_d \\[10pt] y'_d \\[10pt] z'_d \end{bmatrix} =
        \begin{bmatrix}
            \dfrac{ (x_d - c_{d,x}) \cdot z_d }{ f_{d,x} } \\[10pt]
            \dfrac{ (y_d - c_{d,y}) \cdot z_d }{ f_{d,y} } \\[10pt]
            z_d
        \end{bmatrix}
\end{equation}
\begin{equation}
\label{reg_rgb_cam}
    X'_{rgb} = R^{-1} \cdot (X'_d - T)
\end{equation}
\begin{equation}
\label{reg_rgb}
    X_{rgb} = \begin{bmatrix} x_{rgb} \\[10pt] y_{rgb} \\[10pt] z_{rgb} \end{bmatrix} =
        \begin{bmatrix}
            \dfrac{ x'_{rgb} \cdot f_{rgb,x} }{ z'_{rgb} } + c_{rgb,x} \\[10pt]
            \dfrac{ y'_{rgb} \cdot f_{rgb,y} }{ z'_{rgb} } + c_{rgb,y} \\[10pt]
            z'_{rgb}
        \end{bmatrix}
\end{equation}

Points $X_{rgb}$ in (\ref{reg_rgb}) can then be mapped to a 2D $W_{rgb} \times H_{rgb}$ size image, forming a registered depth image.
This process summarizes the standard registration algorithm. However, the computation of registered RGB images is needed for target generation. So the $RGB \mapsto D$ mappings produced by Eqs.~(\ref{reg_depth_cam})--(\ref{reg_rgb}) are reversed to build a 2D $W_{d} \times H_{d}$ image of color values instead. Of course, doing this still leaves depth holes with no RGB value attributed, weakening the whole purpose of performing RGB-D registration. To overcome this, depth holes are filled with pixel interpolation and blurring to create smooth transitions between edges of known depth regions.


\vspace{10pt}\noindent\textbf{Inpainting}:
After completing the RGB-D registration, we employ a color-guided Fast Marching Method (FMM) inpainting algorithm to generate the target frames for training the denoising network. Following the original FMM inpainting technique~\cite{telea2004image}, the algorithm starts by delineating the boundaries of all hole regions within the image. Subsequently, it performs inpainting from the outer pixels of these boundaries inwards, ensuring that all hole regions are filled.

Our FMM inpainting technique combines ideas presented in \cite{telea2004image,liu2012guided,li2021depth} and introduces novel elements that enable better results in consumer depth maps. Specifically, the pixel weighting function (see Eq.~(\ref{weight_eq})) is different from the original FMM inpainting~\cite{telea2004image}. Concretely, we prioritize the distance factor $w_{dst}$ while dropping the factors $w_{lev}$ and $w_{dir}$. Additionally, we include two novel weights: $w_g$, relating to color guidance~\cite{liu2012guided}; and $conf$, a confidence factor as in~\cite{li2021depth}. All these new insights contribute to the following novel functions:
\begin{equation}
\label{weight_eq}
    w(p,q) = w_{dst}^2(p,q) \cdot w_g(p,q) \cdot conf(q)
\end{equation}
\begin{equation}
\label{wdst_eq}
    w_{dst}(p,q) = \dfrac{d_0^2}{\Vert p - q \Vert^2}
\end{equation}
\begin{equation}
\label{wguide_eq}
    w_g(p,q) = \exp \left( - \dfrac{\Vert G(p) - G(q) \Vert^2}{2 \cdot \sigma_g^2} \right)
\end{equation}
\begin{equation}
\label{conf_eq}
    conf(q) = \dfrac{1}{1 + 2 \cdot T_{out}(q)}
\end{equation}
where $d_0$ is the minimum inter-pixel distance, usually 1, $G$ denotes the guiding image, and $\sigma_g^2$ is its standard deviation. Additionally, $T$ is a distance map that stores the distance of each pixel to the closest initial hole patch boundary, 
and $T_{out}$ is a function that zeroes pixels in the set of initial holes $\Omega$ and assigns $T$ to the remaining.

Furthermore, as in GFMM~\cite{liu2012guided}, the pixel inpainting priority takes into account both the distance to the initial hole boundary, given by $T(p)$, and the guidance value of neighboring pixels, so that homogeneous areas are inpainted before other regions more likely to be transitive or edge areas. However, the priority function used in SReD's inpainting for target generation, ${Pr}(p)$, introduces a new normalization variable $T_{max}$, leading to the final equation:
\begin{equation}
\label{priority_eq}
    Pr(p) = (1 - \lambda) \cdot \dfrac{T(p)}{T_{max}} + \lambda \cdot (1 - S_g(p))
\end{equation}
\begin{equation}
\label{guide_factor_eq}
    S_g(p) = \dfrac{1}{\vert N(p) \vert} \cdot \sum \limits_{q \in N(p)} w_g(p,q)
\end{equation}
where, $S_g(p)$ (Eq.~(\ref{guide_factor_eq})) gives the local guide similarity at pixel $p$, $\vert N(p) \vert$ denotes the number of known pixels in the neighborhood of $p$, $T_{max}$ is the greatest value in distance map $T$, and $\lambda$ is a mixing parameter.
(Note: lower $Pr$ values denote greater priority)



\subsection{Denoising Network}

The denoising neural network implemented in SReD adopts a convolutional autoencoder architecture based on the U-Net design~\cite{ronneberger2015unet}. This architecture is primarily influenced by features from MF2F~\cite{dewil2021self}, FastDVDnet~\cite{tassano2020fastdvdnet}, and Noise2Noise~\cite{lehtinen2018noise2noise}. During inference, the network takes as input three sequential depth frames, specifically \(d_{t-2}, d_{t-1}, d_{t}\). Conversely, during training, the input frames are dilated, namely \(d_{t-4}, d_{t-2}, d_{t}\). This setup enables using an inpainted version of frame \(d_{t-1}\) as the target, thereby preventing the network from learning the identity function. The network employs the Mean Absolute Error (MAE or \(L_1\)) loss function to measure the discrepancy between the inpainted target \(d^*_{t-1}\) and the input depth frame \(d_t\). In noisy regions, this approach replicates the effects 
of Noise2Noise~\cite{lehtinen2018noise2noise}, and for depth holes, 
the network 
learns inpainting techniques.

Using a U-Net~\cite{ronneberger2015unet} helps with image denoising. This is because the skip-connections enable passing higher frequency details from the encoding stage to the decoding 
stages via layer concatenation. This propagation allows the network to ``flatten'' noise areas while still preserving sharp image features, such as object contours. Regarding the layer structure, the general layout loosely follows the model presented in Noise2Noise~\cite{lehtinen2018noise2noise}, differing mainly in the number of channels at each network block and the downsampling/upsampling layers. 
Instead of max pooling and 2D upsampling layers, SReD uses 2D convolutions with stride two and transposed 2D convolutions, giving the model more learning flexibility.

Additionally, like in FastDVDnet~\cite{tassano2020fastdvdnet}, the network applies a final residual operation between the input frame $d_{t}$ and the frame resulting from the last convolutional layer in the model $d_{last}$, yielding the depth frame prediction
$ d_{pred} = d_t - d_{last} $.

\section{Evaluation}
\label{sec:eval}

\begin{figure*}[!t]
   \centering
   \includegraphics[width=\linewidth]{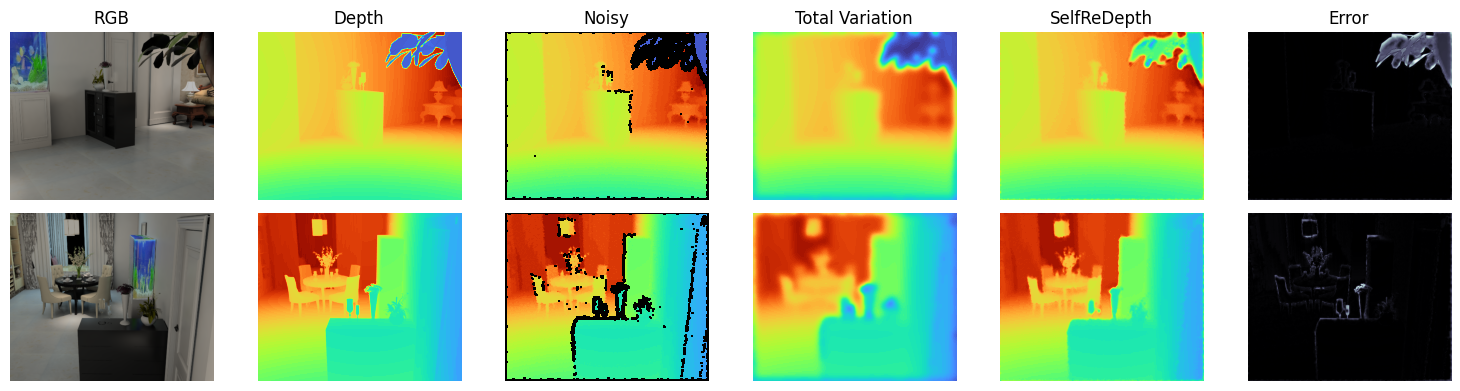}
   \caption{
   Example scenes from the ground-truth dataset demonstrating (from left to right) RGB color image, real depth map, depth map with synthetic noise added
   \textcolor{highlight}{(missing values in black color)},
   results from the Total Variation method, SelfReDepth, and error map from our approach.
   }
   \label{multiCompare}
\end{figure*}

We conducted a comprehensive evaluation of
our method
to assess the algorithm's performance.
A quantitative evaluation
was performed
using a reference-independent noise metric to measure 
SReD's
depth-restoration
capabilities objectively. 
We also performed several tests using a synthetic dataset that provides usable artificial ground-truth data.
The algorithm's time performance 
was also assessed
to determine its suitability for real-time applications.
Furthermore, we evaluated the method's temporal coherence 
using a specialized 
metric.
Finally, we compared SReD 
to other relevant reference-independent
restoration algorithms to situate its performance within the broader landscape of available techniques.

\subsection{Data and Metrics}

Identifying an appropriate combination of data and metrics for evaluating SReD proved to be a non-trivial task. Ideally, we would have access to a consumer-grade depth video dataset featuring raw frame sequences and reference depth data, perhaps captured using a high-precision laser sensor. However, such a dataset is not readily available. This very challenge underscores the importance of developing self-supervised depth denoisers like SReD.

We evaluated SReD on a depth video dataset devoid of reference depth. The evaluation also used appropriate reference-independent metrics. We conducted comprehensive tests on the CoRBS dataset~\cite{wasenmuller2016corbs}, explicitly focusing on the Kinect v2 subset. These data include five distinct RGB-D frame sequences capturing a stationary scene with a mobile camera, resulting in an aggregate of approximately 14,000 depth frames.
In terms of evaluation metrics, the denoised depth frames underwent quantitative assessment concerning noise through a ``non-reference metric for image denoising"~\cite{kong2013new} (NMID),
\textcolor{highlight}{a robust measure based on structure similarity maps from both homogeneous and highly-structured regions, in the absence of the original clean data.}

Additional tests relied on synthetic depth data from the InteriorNet dataset published in \cite{li2018interiornet}, which provides computer-rendered RGB and depth images for various indoor scenes.
For evaluating this synthetic ground-truth data, we employed proper reference metrics for the comparisons: MSE, PSNR, and SSI.
Since this dataset does not provide noisy data, we introduced synthetic noise using the developed
\textcolor{highlight}{Kinect v1 noise model from Handa et al. \cite{handa2014benchmark},
which combines Gaussian noise, bilinear interpolation, and quantization to produce noisier pixels at higher distances and missing depth values at pixels whose corresponding normals are close to perpendicular to the camera's viewing direction.}

To further assess our approach's feasibility 
on these data, we evaluated it against the Total Variation (TV) method \cite{chambolle2004algorithm}.
\textcolor{highlight}{This denoising technique reduces the total magnitude of the image's colour intensity gradient
while simultaneously trying to keep object boundaries.
The regularization parameter
weight used in the algorithmic implementation from \cite{van2014scikit},
which controls the denoising strength at the expense of fidelity to the original image, was set to 0.4 as this value maximized the mean scores among both datasets in our experiments.}

Furthermore, we assess 
temporal coherence using straightforward image differences, 
$M_{temp} = \mathrm{mean}(I_{t+1} - I_{t})$.
We address 
depth value oscillations over time 
by 
analysing granular noise values on a frame-by-frame basis over 
contiguous video sequences from the dataset. 

\begin{figure*}[!h]
    \centering
    \includegraphics[width=0.9\linewidth]{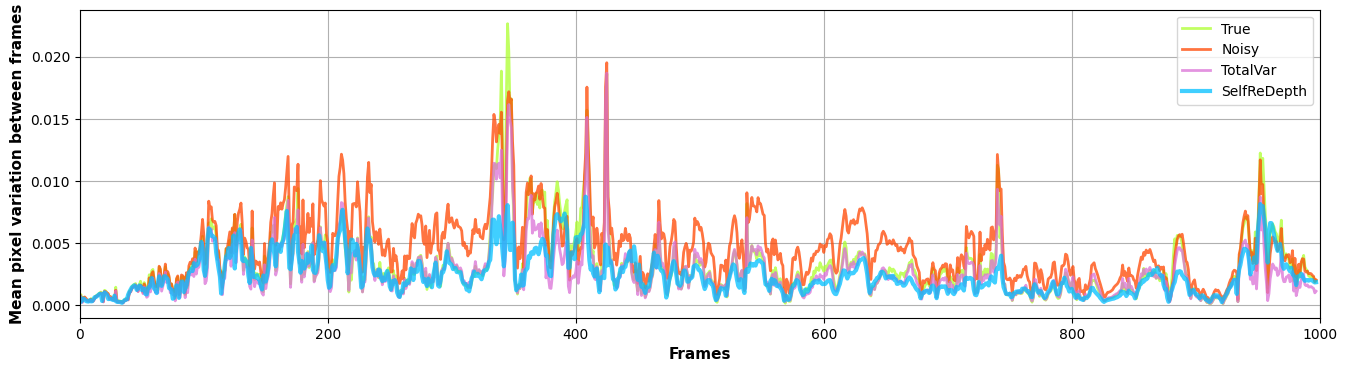}
    \caption{
    \textcolor{highlight}{Temporal analysis of the mean depth value differences across frames. The noisy data were generated from a sample video sequence extracted from the InteriorNet synthetic dataset and restored using the SelfReDepth and Total Variation methods.}
    }
    \label{figTemporal}
\end{figure*}

\subsection{Experimental Setup}

We ran all experiments on a Windows 10 desktop machine with an NVIDIA GeForce RTX 3080 GPU, a Ryzen 7 3700x 8-core CPU, 16 GB of RAM, and an SSD disk drive. We developed and tested SReD
using Python 3.10.8 and tensorflow-gpu 2.10, along with CUDA 11.2 and cuDNN 8.1. 

\subsection{Results}
\label{sec:results}

We trained SReD with batch sizes of 16 and 200 epochs on the CoRBS~\cite{wasenmuller2016corbs} dataset, which 
we also thoroughly shuffled and set with validation and test splits of 0.1 and 0.04, respectively. 

\begin{figure*}[!ht]
    \centering
    \includegraphics[width=0.24\linewidth]{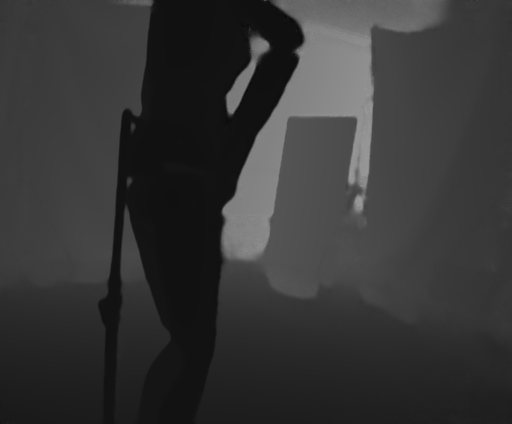}
    \includegraphics[width=0.24\linewidth]{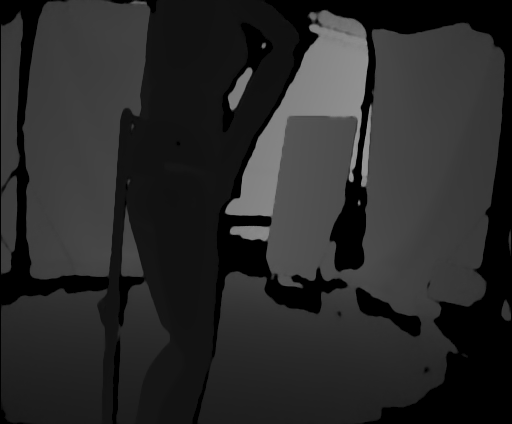}
    \includegraphics[width=0.24\linewidth]{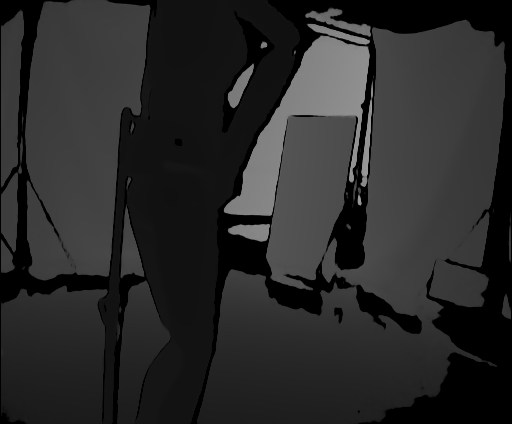}
    \includegraphics[width=0.24\linewidth]{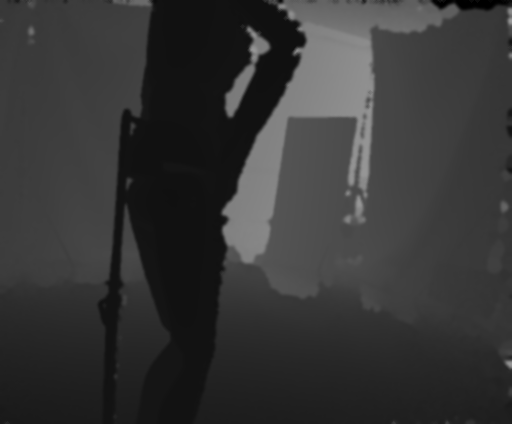}
    \caption{Visual comparison between four image restoration algorithms applied to an example image taken from real data from the CoRBS dataset. From left to right: SelfReDepth, Noise2Noise, Noise2Stack and FMM+BF.
    }
    \label{figrescompare}
\end{figure*}

In the produced denoised depth maps, in Fig.~\ref{figrescompare},
it can be seen that the model learned how to attenuate the noise in the original depth map and fill depth holes. Moreover, on the hardware used for evaluation, the model takes, on average, 9ms to denoise each depth map. Given that a regular RGB-D sensor, such as the Kinect v2, records data at a frequency of 30 frames-per-second (33ms per frame), this evaluation confirms that the model can achieve the desired real-time performance during inference.


We also compared SReD against 
other approaches, including two deep-learning methods,
Noise2Stack~\cite{papkov2021noise2stack} and Noise2Noise~\cite{lehtinen2018noise2noise} and 
two deterministic approaches, the Total Variation method \cite{chambolle2004algorithm} and a combination of a pair of methods that applies FMM inpainting~\cite{telea2004image} 
followed by Bilateral Filtering~\cite{tomasi1998bilateral} denoising. We chose Noise2Noise to evaluate how the implemented technique differs from the original self-supervised U-Net~\cite{ronneberger2015unet} denoiser and what benefits were secured by targeting specifically the denoising of depth. Similarly, we chose Noise2Stack~\cite{papkov2021noise2stack} to compare SReD against another spatio-temporal depth denoiser. 
Last, we used the deterministic FMM+BF combination 
to evaluate how SReD fares against more traditional approaches that perform
\textcolor{highlight}{both denoising and inpainting.}

As already mentioned, we used a non-reference noise metric, NMID~\cite{kong2013new}, 
to quantify the denoising quality, and applied a direct image difference metric 
to evaluate temporal coherence on contiguous depth videos. The results 
can be seen in Table~\ref{metrics}.
From the measured values, 
we note that SReD attained promising results, rivalling the significantly more computationally expensive deterministic algorithms with the NMID metric and achieving the best results with the temporal coherence 
\textcolor{highlight}{in both datasets as expected, since it relies on multiple consecutive frames for its inference process.}
In addition, results in Fig.~\ref{figTemporal} show that 
original per-frame noise discrepancies are mostly fixed, and 
yield temporarily consistent values after denoising.


As shown in Fig.~\ref{figrescompare}, Noise2Noise and Noise2Stack 
can only perform pixel denoising and not depth completion.
As for the deterministic algorithm combination, while it was capable of denoising and inpainting the depth maps, it can also be visually seen that both the edge preservation and depth completion results are inferior to 
those of SReD.


\subsection{Discussion}
\label{sec:discussion}

Based on the experiments and metrics, SReD effectively learned to reduce noise in depth maps. However, some image details still need to be recovered, evident in the blurred features of the doll in the CoRBS~\cite{wasenmuller2016corbs} dataset.
In these cases where large black depth holes are present, inpainting struggles to effectively reconstruct this missing data due to the absence of depth information, resulting in over-smoothed restorations.
While the results are promising, they highlight the need for further work on detail preservation.

\textcolor{highlight}{Additionally, our method seems to struggle with accurately restoring the depth of objects very close to the camera, as seen in the top row of Fig.~\ref{multiCompare} and its rightmost error profile. This scenario where the objects are almost touching the camera was not seen in the training data but is very frequent in the synthetic dataset, signalling the need to extend the training set to a wider range of scenarios.}

The metrics 
pitched SReD promisingly against the other four algorithms evaluated. The visual analysis of the denoised data aligns well with the NMID metric values, reinforcing its reliability.
Although not optimal, our method also performed very favourably in the synthetic ground-truth dataset regarding MSE, PSNR, and SSI scores.
\textcolor{highlight}{The subpar results on this synthetic dataset could be related to the a posteriori added synthetic noise based on a Kinect v1 noise model, while SReD was trained on Kinect v2 noise. This discrepancy, limited by the nonexistence of a usable v2 noise model implementation, might explain the better results achieved in this dataset by more general image restoration approaches not based on deep learning.}

\textcolor{highlight}{In qualitative terms, upon visual inspection, SReD achieved both consistent inpainting and denoising behaviour and
outperforms both deterministic approaches, namely the FMM+BF algorithm, particularly when filling missing areas and sharpening object boundaries, and TV, as this last method over-smooths the overall depth image, failing to preserve object details.}

\subsection{Real-Time Performance}
\label{sec:Real-Time-Performance}

Finally, 
the implemented algorithm can produce denoised frames at frequencies higher than 30 frames per second, thus making SReD suitable for real-time use. Indeed, on the computer hardware used for evaluation, the model requires, on average, 9ms to denoise each depth map. Given that commercial off-the-shelf RGB-D sensors, such as the Kinect v2, generate data at 30 frames-per-second (33ms per frame), our technique can achieve real-time performance at even higher frame rates. 
The modular design of the U-Net architecture allows for straightforward scalability to accommodate larger image sizes without a significant impact on computational time, thereby maintaining real-time performance, as will be detailed in the following section.

\vspace{10pt}\noindent
\textbf{SelfReDepth Complexity}:
\label{sec:SelfReDepth Complexity} 
Our approach is made up of three main blocks as follows: inpainting (eqs. (\ref{FMM_original_full_eq}) and (\ref{weight_eq})), registration  (eqs. (\ref {reg_depth_cam}) - (\ref{reg_rgb})) and denoising (U-Net network) procedures.
As already stated in Sec.~\ref{sec:SelfReDepth-Architecture}, our proposal is designed to satisfy real-time requirements, as we can modify the SReD architecture during the inference stage, which is the one that has a direct impact on the complexity budget. Specifically, (i) remove the target generation, and (ii) the denoise shift to non-dilated input. Thus, only the U-Net (denoiser) needs to be carefully addressed since it is the unique block that affects a constrained time budget requirement.

\begin{table*}[!h]
\caption{
\textcolor{highlight}{Benchmark of several metrics for all the tested methods: NMID (higher is better), temporal difference (lower is better), PSNR (higher is better), MSE and SSI (lower is better). The last three metrics are reported for both real (CoRBS) and synthetic (InteriorNet) datasets while the first two are only available for the synthetic dataset since this is the only one which provides ground-truth depth data.}
}
\label{metrics}
\centering
\begin{tabular}{l|cc|ccccc}
                & \multicolumn{2}{c|}{CoRBS} & \multicolumn{5}{|c}{InteriorNet} \\
                \cline{2-3} \cline{4-8}
                & NMID   & Temporal         & NMID    & Temporal & PSNR   & MSE     & SSI \\
\hline
SelfReDepth     &  \textbf{0.735} & \textbf{0.858} &  0.1611 & \textbf{0.0024}   & 38.663 & 0.00036 & 0.830 \\
Noise2Noise     & -0.154 & 1.085 &  0.1009 & 0.0039   & 39.020 & 0.00013 & 0.937 \\
Noise2Stack     & -0.166 & 1.012 & -0.0504 & 0.0038   & 37.582 & 0.00020 & 0.918 \\
FMM + BF        &  \textbf{0.735} & 0.926 &  \textbf{0.2153} & 0.0032   & \textbf{43.753} & 0.00008 & 0.971 \\
Total Variation &  0.165 & 1.057            &  0.0129 & 0.0026   & 42.979 & \textbf{0.00006} & \textbf{0.980} \\
\end{tabular}
\end{table*}


\vspace{10pt}\noindent\textbf{Time Complexity}: 
We detail the architecture adaptation under a given complexity budget. The choice of the U-Net provides flexibility because it is possible to adapt its architecture under a predefined budget. 
The designs of network architectures should exhibit tradeoffs among several of its components, {\em, i.e.} depth, numbers of filters, and filter sizes, from which the scalability is accomplished. From the above, depth is the most influential concerning the accuracy. Although it is not a straightforward observation, previous work~\cite{simonyan2014very,szegedy2015going} has demonstrated its impact. 
The total time complexity of all the convolutional layers is given as
${\cal O}\Bigl(  \sum_{l=1}^d c_{l-1}\;.\; s_{l}^2\; .\; f_l\; .\; m_{l}^2   \Bigr)$
where $d$ is the depth of the network, ({\em i.e.} the number of convolutional layers), $l$ indexes the convolutional layer, $c_{l}$ is the number of input channels in the $l$-th layer, $f_l$ is the number of filters in the $l$-th layer, ({\em i.e.}, the width), $s_l$  and $m_l$ are the spatial size of the filter and the size of the output of the feature map, respectively.
The time cost of fully connected and pooling layers is not considered since these layers take about 5-10$\%$ computational time. The 
time complexity above
is the basis of the network designs, from which we consider the tradeoffs between the depth $d$ and filter sizes $f_l$, inferring how the network scales in time. 

Concretely, we design a model by replacing the layers in our experimental evaluation. This means that when we replace a few layers with some other layers, we must guarantee that the complexity is preserved without changing the remaining layers in the architecture. 
To design such a replacement, 
we progressively modify the model and observe 
the changes in accuracy. 
Our method addresses the following tradeoffs:
\begin{enumerate}
\item depth $d$ and filter sizes $s_l$,
\item depth $d$ and width $f_l$, and
\item width $f_l$ and filter sizes $s_l$.  
\end{enumerate}
We illustrate one of the steps above, the remaining with an analogous procedure, only changing the corresponding parameters accordingly.  
For instance, as an illustrative of step 1 
(tradeoff between depth $d$ and filter size $s$), we replace a larger filter, say $s_1$, with a cascade of smaller filters, say $s_2$. Denoting the layer configuration as above, 
    $L_{\rm conf} = c_{l-1}\,.\, s_{l}^2\,.\,f_l$  
and considering two instances of filter sizes, {\em e.g.} $s_1=3$
$s_2 =2$, and $c_{l-1}=N$, $f_l=N$, we have the following complexities: 
\begin{align} 
\label{jan-3}
{\cal O}_1  &=   N^2 \;.\; s_1^2  \notag\\ 
{\cal O}_2  &=   2 \;.\;( N^2 \;.\; s_2^2)  \notag 
\end{align}
This replacement 
a $s_1\times s_1$ layer with N input/output channels is replaced by two $s_2\times s_2$ layers with N input/output channels. After the above replacement,
the complexity involved in these layers is nearly unchanged, with the reduction fraction of $2s_1^2 / s_2^2\approx 1$.

With the strategy above, we can ``deepen'' the network under the same complexity time budget. 
This allows us to obtain several architectures and pick the best accuracy.
In concrete, from our experimental evaluation 
yielding the times 
mentioned in Section~\ref{sec:Real-Time-Performance}, we found the best accuracy using the configuration
in our final network,
which has 31 layers, 1729 convolutional size three filters, 
yielding 
1 260 865 trainable parameters.

\vspace{10pt}\noindent\textbf{Scalability}: 
Now, we delve into how the architecture scales with the size of input images.
First, let us  introduce 
some basic notation:
\begin{itemize}
    \item $\rm conv2D_{F,st}$ : 2D (contraction) convolution with $F$ number of filters and with stride $st$
    \item $\rm conv2D^{\top}_{F,st}$ : 2D (expansion) transpose convolution with $F$ number of filters and with stride $st$
\end{itemize}

\noindent Our U-Net network includes the following main blocks:
\begin{itemize}
    \item First Block: 2 x $\rm conv2D_{32,1}$
    \item $i$th Down Block: 1 x $\rm conv2D_{F_i,2}$ + 1 x $\rm conv2D_{F_i,1}$
    \item $i$th Up Block: 2 x $\rm conv2D_{F_i,1}$ + 1 x $\rm conv2D_{F_i,2}^{\top}$
    \item Last Block: 2 x $\rm conv2D_{32,1}$ + 1 x $\rm conv2D_{1,1}$
\end{itemize}
We use five blocks for each Down and Up stage, thus having  $i\in\{1,...,5\}$. The number of filters for each block is
$F = 
\begin{bmatrix}
    F_0 & \dots & F_5
\end{bmatrix}
=
\begin{bmatrix}
    32 & 32 & 48 & 48 & 64 & 128
\end{bmatrix}$, where $F_0$ accounts for the filter in the First and Last blocks.

Now, it is straightforward to determine 
convolutions. Assuming an image size of $W \times H$, we have:
\begin{itemize}
    \item First Block: $2 \cdot F_0 \cdot W \cdot H$
    \item Down Block $i$: $2 \cdot F_i \cdot W \cdot H \cdot 2^{-2i}$
    \item Up Block $i$: $3 \cdot F_i \cdot W \cdot H \cdot 2^{-2i}$
    \item Last Block: $(2 \cdot F_0 + 1) \cdot W \cdot H$
\item \textbf{Total:} $\Bigl(1+4F_0 + 5 \sum_{i=1}^5{F_i \cdot 2^{-2i}}\Bigr) W \cdot H \\ = 189.625 W \cdot  H$    
\end{itemize}

This means that, for a $\Delta$-increment in the image 
resolution $(W+\Delta)(H+\Delta)$, we have a complexity of 
${\cal O}(\Delta^2)$.
So, roughly speaking, a twofold increase in image resolution would entail a fourfold increase in image processing time using the same architecture and memory footprint. Assuming that in the worst case, 80\% of the CPU time is spent on running the Neural Network, the processing time per frame would be around 30ms for an effective frame rate of 30Hz, which is still reasonable.
{\color{highlight}
\section{Limitations}
While our technique has proven to be very effective at restoring depth values from noisy RGB-D images, it can be improved in several ways.
A notable limitation 
involves adequately addressing high-frequency temporal noise. While effective for general noise reduction, 
averaging pixel values across frames falls short in capturing and mitigating these rapid fluctuations. This suggests potential for future refinement. More sophisticated techniques should be capable of discerning and smoothing out high-frequency temporal noise without compromising the dynamic content of the scenes.}

\section{Conclusions and future work}
\label{sec:conclusion}

We introduced SelfReDepth, a self-supervised approach for denoising and completing low-quality depth maps generated from consumer-grade sensors. Our technique advances self-supervised learning in-depth data denoising, offering a precise, data-driven architecture 
without reference data. This flexibility makes SelfReDepth easily adaptable across various environments and applications.

SelfReDepth's architecture features two main elements: a denoising network and a target generation component. The denoising network is inspired by the original Noise2Noise~\cite{lehtinen2018noise2noise} and MF2F~\cite{dewil2021self} video denoisers and is responsible for learning how to denoise depth data without the need for reference data. Meanwhile, the target generation component fills in the gaps in target depth frames using color-guided FMM inpainting. The technique can denoise inaccurate depth values and paint out missing ones with this structure.

We also implemented and assessed SelfReDepth for both denoising efficacy and time performance. Results indicate real-time noise elimination and successful inpainting of depth gaps.
Future work will focus on preserving image details compromised by denoising. Training with synthetic data might also improve depth inpainting performance and dampen oscillations.

\textcolor{highlight}{
In future work, we aim to explore controllable image denoising to generate clean
sample frames with human perceptual priors and balance sharpness and smoothness. In most common 
filter-based denoising
approaches, this can be straightforwardly achieved by regulating the filtering strength. However, for deep neural networks (DNN), regulating the final denoising strength requires performing network inference each time. This of course, hampers the real-time user interaction. Further work will address real-time controllable denoising, to be integrated into a video denoising pipeline that provides a
fully controllable user interface to edit arbitrary denoising levels in real-time with only one-time DNN  inference.
}
SelfReDepth represents a significant advancement in data denoising, tackling noise and depth hole challenges with notable efficiency. The outcomes of our research are encouraging, illustrating the algorithm's capacity to mitigate these problems. However, the concomitant loss of certain image details in the process highlights areas for potential improvement. This observation underscores the need for additional investigation while pointing to clear pathways for refining future algorithm iterations. Such enhancements aim to improve the balance between our denoising algorithm's robustness and critical image detail preservation, enhancing its already remarkable efficiency and making it more applicable to very complex scenarios.

\begin{acknowledgements}
The work reported in this article was partially supported under the auspices of the UNESCO Chair on AI \& VR by national funds through Fundação para a Ciência e a Tecnologia with references
DOI:10.54499/\-UIDB/\-50021/2020,
DOI:10.54499/DL57/2016/CP1368/\-CT0002
and
2022.09212.PTDC (XAVIER project).\\
The SelfReDepth source code is publicly
available at: \\ \url{https://github.com/alexduarte23/sred}
\end{acknowledgements}

\section*{Conflict of interest}
The authors declare that they have no conflict of interest.

\bibliographystyle{spmpsci}      
\bibliography{mybib}   

\begin{thebibliography}{10}
\providecommand{\url}[1]{{#1}}
\providecommand{\urlprefix}{URL }
\expandafter\ifx\csname urlstyle\endcsname\relax
  \providecommand{\doi}[1]{DOI~\discretionary{}{}{}#1}\else
  \providecommand{\doi}{DOI~\discretionary{}{}{}\begingroup
  \urlstyle{rm}\Url}\fi

\bibitem{basso2018robust}
Basso, F., Menegatti, E., Pretto, A.: Robust intrinsic and extrinsic
  calibration of rgb-d cameras.
\newblock IEEE Transactions on Robotics \textbf{34}(5), 1315--1332 (2018)

\bibitem{batson2019noise2self}
Batson, J., Royer, L.: Noise2self: Blind denoising by self-supervision.
\newblock In: Proceedings of the 36th International Conference on Machine
  Learning, pp. 524--533 (2019)

\bibitem{bertalmio2001navier}
Bertalmio, M., Bertozzi, A.L., Sapiro, G.: Navier-stokes, fluid dynamics, and
  image and video inpainting.
\newblock In: Proceedings of the IEEE Conference on Computer Vision and Pattern
  Recognition, vol.~1, pp. I--I (2001)

\bibitem{calvarons2021improved}
Calvarons, A.F.: Improved noise2noise denoising with limited data.
\newblock In: IEEE/CVF Conf. on Computer Vision and Pattern Recognition
  Workshops, pp. 796--805 (2021)

\bibitem{capecci2016physical}
Capecci, M., Ceravolo, M.G., Ferracuti, F., Iarlori, S., Kyrki, V., Longhi, S.,
  Romeo, L., Verdini, F.: Physical rehabilitation exercises assessment based on
  hidden semi-markov model by kinect v2.
\newblock In: IEEE-EMBS International Conference on Biomedical and Health
  Informatics, pp. 256--259 (2016)

\bibitem{cha2019gan2gan}
Cha, S., Park, T., Kim, B., Baek, J., Moon, T.: Gan2gan: Generative noise
  learning for blind denoising with single noisy images.
\newblock arXiv preprint:1905.10488  (2019)

\bibitem{chambolle2004algorithm}
Chambolle, A.: An algorithm for total variation minimization and applications.
\newblock Journal of Mathematical imaging and vision \textbf{20}, 89--97 (2004)

\bibitem{Matterport3D}
Chang, A., Dai, A., Funkhouser, T., Halber, M., Niessner, M., Savva, M., Song,
  S., Zeng, A., Zhang, Y.: Matterport3d: Learning from rgb-d data in indoor
  environments.
\newblock 2017 International Conference on 3D Vision  (2017)

\bibitem{chaudhary2016approach}
Chaudhary, R., Dasgupta, H.: An approach for noise removal on depth images.
\newblock arXiv preprint:1602.05168  (2016)

\bibitem{chen2012depth}
Chen, L., Lin, H., Li, S.: Depth image enhancement for kinect using region
  growing and bilateral filter.
\newblock In: ICPR2012, pp. 3070--3073 (2012)

\bibitem{choi2021selfdeco}
Choi, J., Jung, D., Lee, Y., Kim, D., Manocha, D., Lee, D.: Selfdeco:
  Self-supervised monocular depth completion in challenging indoor
  environments.
\newblock In: IEEE Int. Conference on Robotics and Automation, pp. 467--474
  (2021)

\bibitem{dai2019new}
Dai, Y., Fu, Y., Li, B., Zhang, X., Yu, T., Wang, W.: A new filtering system
  for using a consumer depth camera at close range.
\newblock Sensors \textbf{19}(16), 3460 (2019)

\bibitem{dewil2021self}
Dewil, V., Anger, J., Davy, A., Ehret, T., Facciolo, G., Arias, P.:
  Self-supervised training for blind multi-frame video denoising.
\newblock In: IEEE Winter Conference on Applications of Computer Vision, pp.
  2724--2734 (2021)

\bibitem{essmaeel2012temporal}
Essmaeel, K., Gallo, L., Damiani, E., De~Pietro, G., Dipanda, A.: Temporal
  denoising of kinect depth data.
\newblock In: Eighth Intl. Conference on Signal Image Technology and Internet
  Based Systems, pp. 47--52. IEEE (2012)

\bibitem{feng2018towards}
Feng, D., Rosenbaum, L., Dietmayer, K.: Towards safe autonomous driving:
  Capture uncertainty in the deep neural network for lidar 3d vehicle
  detection.
\newblock In: 2018 21st International Conference on Intelligent Transportation
  Systems, pp. 3266--3273 (2018)

\bibitem{feng2021advancing}
Feng, Z., Jing, L., Yin, P., Tian, Y., Li, B.: Advancing self-supervised
  monocular depth learning with sparse lidar.
\newblock arXiv preprint:2109.09628  (2021)

\bibitem{gabel2012full}
Gabel, M., Gilad-Bachrach, R., Renshaw, E., Schuster, A.: Full body gait
  analysis with kinect.
\newblock In: Annual International Conference of the IEEE Engineering in
  Medicine and Biology Society, pp. 1964--1967 (2012)

\bibitem{gao2015leveraging}
Gao, Z., Yu, Y., Zhou, Y., Du, S.: Leveraging two kinect sensors for accurate
  full-body motion capture.
\newblock Sensors \textbf{15}(9), 24297--24317 (2015)

\bibitem{handa2014benchmark}
Handa, A., Whelan, T., McDonald, J., Davison, A.J.: A benchmark for rgb-d
  visual odometry, 3d reconstruction and slam.
\newblock In: IEEE international conference on Robotics and automation, pp.
  1524--1531. IEEE (2014)

\bibitem{jiang2018kinect}
Jiang, L., Xiao, S., He, C.: Kinect depth map inpainting using a multi-scale
  deep convolutional neural network.
\newblock In: Proceedings of the 2018 International Conference on Image and
  Graphics Processing, pp. 91--–95 (2018)

\bibitem{jorge19}
Jorge, J., Anjos, R.K.D., Silva, R.: Dynamic occlusion handling for real-time
  ar applications.
\newblock In: Proceedings of the 17th International Conference on
  Virtual-Reality Continuum and Its Applications in Industry (2019)

\bibitem{kong2013new}
Kong, X., Li, K., Yang, Q., Wenyin, L., Yang, M.H.: A new image quality metric
  for image auto-denoising.
\newblock In: IEEE International Conference on Computer Vision, pp. 2888--2895
  (2013)

\bibitem{krull2019noise2void}
Krull, A., Buchholz, T.O., Jug, F.: Noise2void - learning denoising from single
  noisy images.
\newblock In: IEEE/CVF Conference on Computer Vision and Pattern Recognition,
  pp. 2124--2132 (2019)

\bibitem{krull2020probabilistic}
Krull, A., Vi{\v{c}}ar, T., Prakash, M., Lalit, M., Jug, F.: Probabilistic
  noise2void: Unsupervised content-aware denoising.
\newblock Frontiers in Computer Science \textbf{2}, 5 (2020)

\bibitem{kweon2014noise}
Kweon, I.S., Jung, J., Lee, J.Y.: Noise aware depth denoising for a
  time-of-flight camera.
\newblock In: 20th Korea-Japan Joint Workshop on Frontiers of Computer Vision
  (2014)

\bibitem{lachat2015assessment}
Lachat, E., Macher, H., Landes, T., Grussenmeyer, P.: Assessment and
  calibration of a rgb-d camera (kinect v2 sensor) towards a potential use for
  close-range 3d modeling.
\newblock Remote Sensing \textbf{7}(10), 13070--13097 (2015)

\bibitem{lehtinen2018noise2noise}
Lehtinen, J., Munkberg, J., Hasselgren, J., Laine, S., Karras, T., Aittala, M.,
  Aila, T.: Noise2noise: Learning image restoration without clean data.
\newblock In: International Conference on Machine Learning, pp. 2965--2974.
  PMLR (2018)

\bibitem{lemarchand2020noisebreaker}
Lemarchand, F., Findeli, T., Nogues, E., Pelcat, M.: Noisebreaker: Gradual
  image denoising guided by noise analysis.
\newblock In: IEEE 22nd International Workshop on Multimedia Signal Processing,
  pp. 1--6 (2020)

\bibitem{li2021fastcompletion}
Li, A., Yuan, Z., Ling, Y., Chit, W., Zhang, S., Zhang, C.: Fastcompletion: A
  cascade network with multiscale group-fused inputs for real-time depth
  completion.
\newblock In: 25th International Conference on Pattern Recognition, pp.
  866--872 (2021)

\bibitem{li2021depth}
Li, L., Wu, H., Chen, Z.: Depth image restoration method based on improved fmm
  algorithm.
\newblock In: 2021 13th International Conference on Machine Learning and
  Computing, ICMLC 2021, pp. 349--355 (2021)

\bibitem{li2018interiornet}
Li, W., Saeedi, S., McCormac, J., Clark, R., Tzoumanikas, D., Ye, Q., Huang,
  Y., Tang, R., Leutenegger, S.: Interiornet: Mega-scale multi-sensor
  photo-realistic indoor scenes dataset.
\newblock arXiv preprint:1809.00716  (2018)

\bibitem{liu2018image}
Liu, G., Reda, F.A., Shih, K.J., Wang, T.C., Tao, A., Catanzaro, B.: Image
  inpainting for irregular holes using partial convolutions.
\newblock In: Computer Vision -- ECCV 2018, pp. 89--105 (2018)

\bibitem{liu2012guided}
Liu, J., Gong, X., Liu, J.: Guided inpainting and filtering for kinect depth
  maps.
\newblock In: Proceedings of the 21st Int. Conference on Pattern Recognition,
  pp. 2055--2058 (2012)

\bibitem{liu2015detecting}
Liu, J., Liu, Y., Zhang, G., Zhu, P., Chen, Y.Q.: Detecting and tracking people
  in real time with rgb-d camera.
\newblock Pattern Recognition Letters \textbf{53}, 16--23 (2015)

\bibitem{liu2016computationally}
Liu, S., Chen, C., Kehtarnavaz, N.: A computationally efficient denoising and
  hole-filling method for depth image enhancement.
\newblock In: Real-time image and video processing 2016, vol. 9897, pp. 235 --
  243. SPIE (2016)

\bibitem{ma2019self}
Ma, F., Cavalheiro, G.V., Karaman, S.: Self-supervised sparse-to-dense:
  Self-supervised depth completion from lidar and monocular camera.
\newblock In: International Conference on Robotics and Automation, pp.
  3288--3295 (2019)

\bibitem{maimone2012enhanced}
Maimone, A., Bidwell, J., Peng, K., Fuchs, H.: Enhanced personal
  autostereoscopic telepresence system using commodity depth cameras.
\newblock Computers \& Graphics \textbf{36}(7), 791--807 (2012)

\bibitem{mallick2014characterizations}
Mallick, T., Das, P.P., Majumdar, A.K.: Characterizations of noise in kinect
  depth images: A review.
\newblock IEEE Sensors journal \textbf{14}(6), 1731--1740 (2014)

\bibitem{metzler2018unsupervised}
Metzler, C.A., Mousavi, A., Heckel, R., Baraniuk, R.G.: Unsupervised learning
  with stein's unbiased risk estimator.
\newblock arXiv preprint:1805.10531  (2018)

\bibitem{mohan2021adaptive}
Mohan, S., Vincent, J.L., Manzorro, R., Crozier, P., Fernandez-Granda, C.,
  Simoncelli, E.P.: Adaptive denoising via gaintuning.
\newblock In: Thirty-Fifth Conference on Neural Information Processing Systems
  (2021)

\bibitem{moran2020noisier2noise}
Moran, N., Schmidt, D., Zhong, Y., Coady, P.: Noisier2noise: Learning to
  denoise from unpaired noisy data.
\newblock In: IEEE/CVF Conference on Computer Vision and Pattern Recognition,
  pp. 12061--12069 (2020)

\bibitem{newcombe2011kinectfusion}
Newcombe, R.A., Izadi, S., Hilliges, O., Molyneaux, D., Kim, D., Davison, A.J.,
  Kohi, P., Shotton, J., Hodges, S., Fitzgibbon, A.: Kinectfusion: Real-time
  dense surface mapping and tracking.
\newblock In: 10th international symposium on mixed and augmented reality, pp.
  127--136 (2011)

\bibitem{oyedotun2017facial}
Oyedotun, O.K., Demisse, G., El~Rahman~Shabayek, A., Aouada, D., Ottersten, B.:
  Facial expression recognition via joint deep learning of rgb-depth map latent
  representations.
\newblock In: IEEE International Conference on Computer Vision Workshops, pp.
  3161--3168 (2017)

\bibitem{papkov2021noise2stack}
Papkov, M., Roberts, K., Madissoon, L.A., Shilts, J., Bayraktar, O., Fishman,
  D., Palo, K., Parts, L.: Noise2stack: Improving image restoration by learning
  from volumetric data.
\newblock In: Intl. Workshop Machine Learning for Medical Image Reconstruction,
  pp. 99--108 (2021)

\bibitem{quan2020self2self}
Quan, Y., Chen, M., Pang, T., Ji, H.: Self2self with dropout: Learning
  self-supervised denoising from single image.
\newblock In: IEEE/CVF Conference on Computer Vision and Pattern Recognition,
  pp. 1887--1895 (2020)

\bibitem{ren2011robust}
Ren, Z., Yuan, J., Zhang, Z.: Robust hand gesture recognition based on
  finger-earth mover's distance with a commodity depth camera.
\newblock In: Proceedings of the 19th international conference on Multimedia,
  pp. 1093--1096 (2011)

\bibitem{ronneberger2015unet}
Ronneberger, O., Fischer, P., Brox, T.: U-net: Convolutional networks for
  biomedical image segmentation.
\newblock In: International Conference on Medical image computing and
  computer-assisted intervention, pp. 234--241 (2015)

\bibitem{sheth2021unsupervised}
Sheth, D.Y., Mohan, S., Vincent, J.L., Manzorro, R., Crozier, P.A., Khapra,
  M.M., Simoncelli, E.P., Fernandez-Granda, C.: Unsupervised deep video
  denoising.
\newblock In: Proceedings of the IEEE/CVF International Conference on Computer
  Vision, pp. 1759--1768 (2021)

\bibitem{simonyan2014very}
Simonyan, K., Zisserman, A.: Very deep convolutional networks for large-scale
  image recognition.
\newblock arXiv preprint:1409.1556  (2014)

\bibitem{song2017depth}
Song, W., Le, A.V., Yun, S., Jung, S.W., Won, C.S.: Depth completion for kinect
  v2 sensor.
\newblock Multimedia Tools and Applications \textbf{76}(3), 4357--4380 (2017)

\bibitem{szegedy2015going}
Szegedy, C., Liu, W., Jia, Y., Sermanet, P., Reed, S., Anguelov, D., Erhan, D.,
  Vanhoucke, V., Rabinovich, A.: Going deeper with convolutions.
\newblock In: Proceedings of the IEEE conference on computer vision and pattern
  recognition, pp. 1--9 (2015)

\bibitem{tassano2020fastdvdnet}
Tassano, M., Delon, J., Veit, T.: Fastdvdnet: Towards real-time deep video
  denoising without flow estimation.
\newblock In: IEEE/CVF Conference on Computer Vision and Pattern Recognition,
  pp. 1354--1363 (2020)

\bibitem{telea2004image}
Telea, A.: An image inpainting technique based on the fast marching method.
\newblock Journal of Graphics Tools \textbf{9}(1), 23--34 (2004)

\bibitem{tolgyessy2021evaluation}
T{\"o}lgyessy, M., Dekan, M., Chovanec, L., Hubinsk{\`y}, P.: Evaluation of the
  azure kinect and its comparison to kinect v1 and kinect v2.
\newblock Sensors \textbf{21}(2), 413 (2021)

\bibitem{tomasi1998bilateral}
Tomasi, C., Manduchi, R.: Bilateral filtering for gray and color images.
\newblock In: 6th international conference on computer vision (IEEE Cat. No.
  98CH36271), pp. 839--846 (1998)

\bibitem{van2014scikit}
Van~der Walt, S., Sch{\"o}nberger, J.L., Nunez-Iglesias, J., Boulogne, F.,
  Warner, J.D., Yager, N., Gouillart, E., Yu, T.: scikit-image: image
  processing in python.
\newblock PeerJ \textbf{2}, e453 (2014)

\bibitem{wan2020edge}
Wan, Y., Li, Y., Jiang, J., Xu, B.: Edge voxel erosion for noise removal in 3d
  point clouds collected by kinect{\copyright}.
\newblock In: Proceedings of the 2020 2nd International Conference on Image,
  Video and Signal Processing, pp. 59--63 (2020)

\bibitem{wasenmuller2016corbs}
Wasenm{\"u}ller, O., Meyer, M., Stricker, D.: Corbs: Comprehensive rgb-d
  benchmark for slam using kinect v2.
\newblock In: IEEE Winter Conference on Applications of Computer Vision, pp.
  1--7 (2016)

\bibitem{xiong2019foreground}
Xiong, W., Yu, J., Lin, Z., Yang, J., Lu, X., Barnes, C., Luo, J.:
  Foreground-aware image inpainting.
\newblock In: IEEE/CVF Conference on Computer Vision and Pattern Recognition,
  pp. 5833--5841 (2019)

\bibitem{yu2018generative}
Yu, J., Lin, Z., Yang, J., Shen, X., Lu, X., Huang, T.S.: Generative image
  inpainting with contextual attention.
\newblock In: IEEE/CVF Conference on Computer Vision and Pattern Recognition,
  pp. 5505--5514 (2018)

\bibitem{zennaro2015performance}
Zennaro, S., Munaro, M., Milani, S., Zanuttigh, P., Bernardi, A., Ghidoni, S.,
  Menegatti, E.: Performance evaluation of the 1st and 2nd generation kinect
  for multimedia applications.
\newblock In: IEEE International Conference on Multimedia and Expo, pp. 1--6
  (2015)

\bibitem{zhang2008adaptive}
Zhang, B., Allebach, J.P.: Adaptive bilateral filter for sharpness enhancement
  and noise removal.
\newblock IEEE International Conference on Image Processing \textbf{4},
  417--420 (2007)

\bibitem{zhang2012water}
Zhang, X., Yan, J., Feng, S., Lei, Z., Yi, D., Li, S.Z.: Water filling:
  Unsupervised people counting via vertical kinect sensor.
\newblock In: IEEE 9th intl. conference on advanced video and signal-based
  surveillance, pp. 215--220 (2012)

\bibitem{zhou2012study}
Zhou, X.: A study of microsoft kinect calibration.
\newblock Dept. of Comp. Science, George Mason University, Fairfax  (2012)

\end{thebibliography}

\end{document}